\documentclass[runningheads]{llncs}
\usepackage[T1]{fontenc}
\usepackage{graphicx,verbatim}
\usepackage{booktabs}

\usepackage{amsmath}
\usepackage{amsfonts}
\usepackage{algorithm}
\usepackage{algpseudocode} 

\usepackage{booktabs}
\usepackage{multirow}
\usepackage{makecell}
\usepackage[table]{xcolor}
\definecolor{Gray}{gray}{0.9}
\definecolor{lightskyblue}{RGB}{198,226,255}
\usepackage{hyperref}

\usepackage{subcaption} 

\usepackage{bm}        
\usepackage{mathtools} 

\begin{document}
%
\title{From Point Estimates to Distributions: GMM Pooling for MIL in Preterm Birth Prediction}
\titlerunning{GMM-Pooling for MIL in Preterm Birth Prediction}
%
\author{Hussain Alasmawi \inst{1}\and
Numan Saeed \inst{1} \and
Soha Said \inst{2} \and 
Mohammad Yaqub \inst{1} }
%
\authorrunning{H. Alasmawi et al.}
%
\institute{Division of Computing and Mathematical Sciences, Mohamed bin Zayed University of Artificial Intelligence (MBZUAI), Abu Dhabi, United Arab Emirates
 \and
Corniche Hospital, Abu Dhabi Health Services Company (SEHA), Abu
Dhabi, UAE\\
\email{hussain.alasmawi@mbzuai.ac.ae}}


  
\maketitle              
\begin{abstract}

Preterm birth (PTB) prediction can enable targeted surveillance and timely intervention, yet most ultrasound-based models use a single selected transvaginal ultrasound (TVUS) frame per patient despite routine exams acquiring multiple cervical images. We formulate PTB prediction as a multiple instance learning (MIL) problem, representing each patient as a variable-sized bag of TVUS images with a single outcome label. To move beyond standard MIL aggregators that collapse a bag into a point estimate, we propose a Gaussian Mixture Model (GMM) pooling, which summarizes all images in a bag into a fixed-length representation by modeling their feature distribution. This design captures intra-patient variability. We evaluate the method on a private clinical cohort and on a public lymph node metastasis benchmark. For PTB prediction, GMM pooling improves over the instance-based model PR-AUC from 0.44  to 0.56. On the lymph node benchmark, it achieves state-of-the-art performance with 0.91 F1-score and 0.89 ROC-AUC for classification and 0.18 MAE for regression. The code is publicly available at \textbf{\url{https://github.com/HussainAlasmawi/GMM_Pooling}}.



\keywords{multiple instance learning \and preterm birth \and distribution pooling \and ultrasound}
\end{abstract}

 \section{Introduction}
 
Preterm birth (PTB), defined as delivery before 37 weeks of gestation, remains a major contributor to neonatal morbidity and mortality worldwide and affects approximately 10\% of pregnancies \cite{WHO2023preterm}. Importantly, three-quarters of preterm-related deaths could be prevented with timely and appropriate medical management \cite{behrman2007preterm}. Early risk stratification can inform preventive therapy, antenatal surveillance, referral, and neonatal readiness.  In routine practice, PTB risk assessment relies on clinical factors and TVUS, most notably measuring cervical length as a biomarker of PTB. However, cervical length measurement is operator-dependent and exhibits inter- and intra-observer variability in cervical view acquisition \cite{Gravett2024}.   Reflecting this uncertainty, clinical guidelines  (e.g., ISUOG) recommend acquiring multiple cervical images and reporting the shortest valid measurement \cite{ISUOG2022preterm}.

Recent machine learning work has explored PTB prediction from electronic health records \cite{huang2024ehr,kloska2025ehr,gao2019ehr,koivu2020predicting} or ultrasound imaging \cite{wlodarczyk2019estimation,wlodarczyk2020spontaneous,ohtaka2024deep}. More broadly, deep learning-based fetal health assessment from ultrasound imaging has been extensively studied for tasks such as congenital heart disease detection \cite{arnaout2021ensemble,taratynova2025tpa}, synthetic fetal view generation \cite{tian2025enhancing,arjemandi2025difusal}, and ultrasound view classification \cite{baumgartner2016real,maani2025fetalclip} or clustering \cite{alasmawi2024fusc}. In contrast, comparatively less attention has been given to the prediction of preterm birth from ultrasound data. 
Moreover, existing PTB image-based approaches generally operate on a single selected image per patient, even though multiple images are often acquired during a routine exam. This selection can discard informative within-exam variability (e.g., view differences and acquisition conditions) and may miss subtle cues that appear only in a subset of images, limiting PTB prediction. While cervical length is the standard ultrasound biomarker for PTB risk assessment, it summarizes the examination using a single measurement, may not fully capture the information contained across multiple acquired images, and has shown limited predictive power for PTB prediction \cite{conde2015predictive}.

To leverage all available images under patient-level supervision, we formulate PTB prediction as a multiple instance learning (MIL) problem, where each patient corresponds to a bag of TVUS images with a single outcome label. A fundamental challenge in MIL is aggregating a variable number of instance-level features to produce a fixed-dimensional representation while ensuring permutation invariance to instance ordering. Standard pooling modules (max, mean, attention \cite{ilse2018attention}) summarize a bag as a single point estimate in feature space. While effective, point-estimate pooling can create an information bottleneck by collapsing within-bag variability that may be informative for bag-level prediction \cite{oner2023distribution}. This motivates distributional pooling approaches that aim to preserve richer information beyond a single summary vector.

Distribution-based MIL pooling, which models feature-wise marginal densities, has been proposed to retain richer information by representing a bag through estimated feature distributions \cite{oner2023distribution}. In this work, we refer to such approaches as density-based pooling. In these formulations, the bag representation is constructed by modeling each feature dimension independently. While this strategy captures variability beyond point-estimate pooling (e.g., max or mean), it does not explicitly account for cross-dimensional dependencies within the feature space. To address this limitation, we propose a \textbf{ Gaussian mixture model (GMM) pooling}, which models the within-bag feature distribution using a mixture model estimated end-to-end. The method learns (i) soft instance-to-component responsibilities and (ii) instance importance weights, and then produces a fixed-length bag embedding by evaluating the learned mixture density at a set of learnable probe vectors.  


In summary, our contributions are: (i) a GMM pooling module for MIL that
preserves informative within-bag feature distributions, addressing the
limitations of point-estimate pooling; (ii) a differentiable distribution embedding
constructed via log-density evaluations at learnable probe vectors; and (iii) a
comprehensive empirical evaluation on a transvaginal cervical ultrasound cohort
and a public lymph node metastasis MIL benchmark, demonstrating that MIL improves over instance-level prediction and that GMM pooling is competitive across classification and regression tasks.


\begin{figure}[h]
    \centering
    \includegraphics[width=1\textwidth]{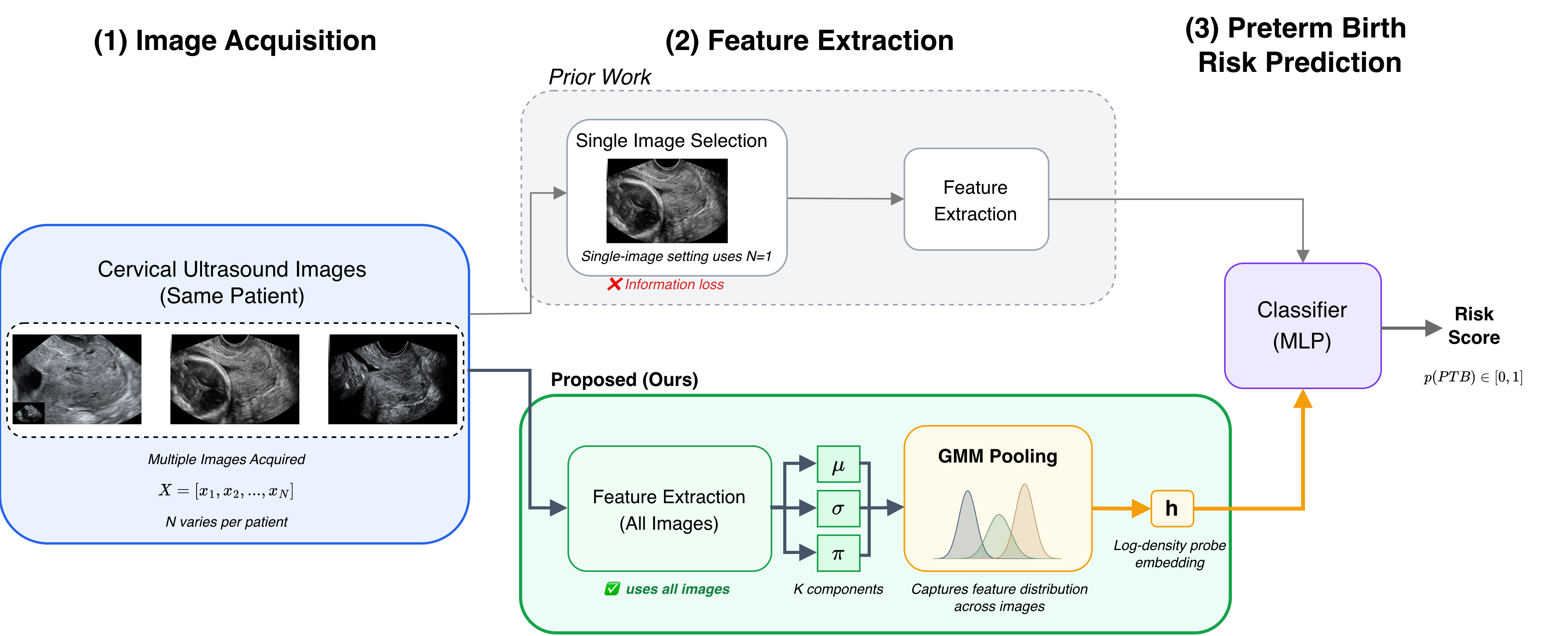}
    \caption{Clinicians acquire multiple cervical ultrasound images per patient; prior work uses only one image, whereas we leverage all images using MIL with GMM pooling to capture feature distributions for PTB prediction. Here, $\mu$, $\sigma$, and $\pi$ denote the mean, (diagonal) standard deviation, and mixing weight of each Gaussian component, respectively; $K$ is the number of mixture components; and $\mathbf{h}$ is the log-density probe embedding computed from the GMM and used for risk prediction.}

    \label{fig:motivation_figure}
    
\end{figure}
\section{Methodology}
We model PTB prediction from TVUS as an MIL problem. Each patient corresponds to a bag of $N$ cervical ultrasound images $X = \{x_1, …, x_N\}$ with a single bag-level label $Y$ indicating the delivery outcome (PTB vs. term). The overall pipeline is illustrated in Figure \ref{fig:motivation_figure}: we extract per-image features, aggregate them into a patient-level representation using GMM pooling, to predict the PTB risk.

\subsection{Problem Formulation and MIL Framework}

\begin{figure}[h]
    \centering
    \includegraphics[width=1\textwidth]{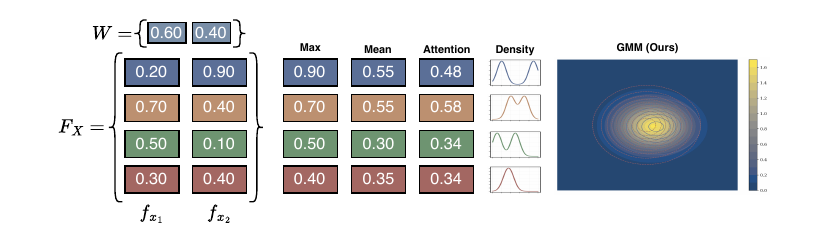}
    \caption{
Illustration of different MIL pooling strategies on a toy example. 
Given instance features $F_X$, max and mean pooling produce a single point-estimate representation. 
Attention pooling computes a weighted point estimate using learned instance importance weights $W$, but still collapses the bag into a single point-estimate representation.  
Density pooling models each feature dimension independently, whereas GMM pooling preserves the joint distributional structure across dimensions.
}

    \label{fig:pooling_example}
    
\end{figure}

Given a bag $X \subseteq \mathcal{I}$ from the instance space $I$ and a bag label $Y \in \mathcal{Y}$, MIL aims to learn a function $f$ such that $\hat{Y} = f(X)$. We propose a standard three-stage architecture:
\begin{enumerate}
     
 \item Feature extraction. A shared encoder $\theta_{feature} : I \rightarrow \mathbb{R}^J$ maps each image $x_i$ to an instance feature vector $f_{x_i} \in \mathbb{R}^J$. Stacking features yields $F_X = [f_{x_1}, …, f_{x_N}] \in \mathbb{R}^{J \times N}$. In the PTB experiments, the feature extractor is implemented using a U-Net \cite{ronneberger2015u} backbone, jointly trained for cervical canal segmentation and bag-level classification. The segmentation task serves as an auxiliary objective that encourages the encoder to learn anatomically meaningful cervical representations, providing an inductive bias for the shared feature space used for bag-level classification. The segmentation branch predicts the cervical mask, while intermediate encoder features serve as instance representations for MIL aggregation.

\item Bag aggregation. A pooling module $\theta_{filter}: \mathbb{R}^{J \times N} \rightarrow H$ aggregates instance features into a fixed-dimensional bag representation $h \in \mathcal{H}$. The aggregation must be permutation-invariant and able to handle variable bag sizes. Figure \ref{fig:pooling_example} contrasts common MIL pooling modules (max/mean/attention \cite{ilse2018attention} and density-based \cite{oner2023distribution} pooling) with our GMM pooling.

\item Prediction. A head $\theta_{head}: \mathcal{H} \rightarrow \mathcal{Y}$ maps $h$ to the predicted bag label $\hat{Y}$.
\end{enumerate}

\subsection{ Gaussian Mixture Model Pooling}

Standard pooling modules $\theta_{\text{filter}}$ reduce a bag to a point estimate in feature space, which can collapse informative within-bag variability. We instead estimate a GMM over instance features and embed the resulting distribution into a fixed-length vector. Algorithm \ref{alg:gmm_pooling} summarizes the full method.

Given $F_X \in \mathbb{R}^{J \times N}$, we construct a GMM with $M$ components. Each component is parameterized by a mixing coefficient $\pi_k$, mean $\mu_k \in \mathbb{R}^J$, and diagonal variance $\sigma_k^2 \in \mathbb{R}^J$. We adopt a diagonal covariance parameterization for computational stability and scalability, as optimizing full covariance matrices within an end-to-end MIL framework would substantially increase the number of parameters and complicate training. We did not explore full or low-rank in this work. Nevertheless, soft instance-to-component assignments and mixture modeling preserve heterogeneous intra-bag structure and relationships among instances. All parameters are estimated from the bag features in a differentiable manner.  


\paragraph{Soft instance-to-component assignments.}
We obtain soft assignments of instances to mixture components using a responsibility network $\mathbf{f_r} : \mathbb{R}^{J} \rightarrow \mathbb{R}^{\mathcal{M}}$. 
For each instance feature ${f}_{x_i}$, the network produces a probability distribution over components:
\begin{equation}
r_{i,k} = \mathrm{softmax}_k \big( \mathbf{f_r}(f_{x_i}) \big),
\end{equation}
where $r_{i,k}$ denotes the responsibility of instance $i$ for component $k$, satisfying  $\sum_{k=1}^{\mathcal{M}} r_{i,k} = 1$. Thus, each instance contributes softly to multiple components. In practice, $\mathbf{f_r}$  is implemented as a lightweight two-hidden-layer multilayer perceptron MLP with ReLU activations,
applied independently to each instance feature.

\paragraph{Instance importance weighting.}

In addition to instance responsibilities, we estimate the relative importance
of instances within a bag using an instance importance network
$\mathbf{f_w}: \mathbb{R}^{J} \rightarrow \mathbb{R}$.
Instance weights are computed via a softmax over instances:
\begin{equation}
w_i = \mathrm{softmax}_i \big( \mathbf{f_w}(f_{x_i}) \big),
\end{equation}
ensuring $\sum_{i=1}^{N} w_i = 1$.
Instance importance weighting adjusts how strongly each instance contributes
to the estimated distribution, without enforcing hard instance selection. In practice, $\mathbf{f_w}$ is a lightweight single-hidden-layer MLP applied independently to each instance feature.

\paragraph{ GMM parameter estimation.}

The final attention of instance $i$ to component $k$ is obtained by combining
instance importance and component responsibility:
\begin{equation}
a_{i,k} = w_i \cdot r_{i,k}.
\end{equation}


Using the attention weights $a_{i,k}$, the GMM parameters are estimated as
\begin{equation}
\pi_k = \sum_{i=1}^{N} a_{i,k},
\qquad
\mu_k = \frac{\sum_{i=1}^{N} a_{i,k} f_{x_i}}{\sum_{i=1}^{N} a_{i,k}},
\qquad
\sigma_k^2 =
\frac{\sum_{i=1}^{N} a_{i,k}\left(f_{x_i}-\mu_k\right)^2}
{\sum_{i=1}^{N} a_{i,k}} .
\end{equation}
The mixing coefficients $\pi_k$ are subsequently normalized across components to form a valid probability distribution. The square is applied element-wise, and all parameters are learned end-to-end.

\paragraph{Distribution embedding via learned probes.}
To produce a fixed-dimensional bag representation, we evaluate the log-density of the estimated GMM on a set of $\mathcal{P}$ learnable probe vectors $\{\mathbf{z}_p \in \mathbb{R}^{J}\}_{p=1}^{\mathcal{P}}$. For each probe, we compute
\begin{equation}
h_p =
\log \left(
\sum_{k=1}^{\mathcal{M}}
\pi_k \,
\mathcal{N}({z}_p \mid
{\mu}_k, {\sigma}_k^2)
\right)/{(J \times T)}.
\end{equation}
where division by $J \times T$ controls the embedding scale. The resulting vector $\mathbf{h} = [h_1, \ldots, h_{\mathcal{P}}] \in \mathbb{R}^{\mathcal{P}}$
serves as the distribution-based bag embedding as the output of the pooling module $\theta_{\text{filter}}$.

\begin{algorithm}[th]
\caption{GMM Pooling ($\theta_{\text{filter}}$)}
\label{alg:gmm_pooling}
\scriptsize 
\begin{algorithmic}[1]
\Statex \textbf{Input:} Instance features $ F_X = [f_{x_1}, …, f_{x_N}]\in\mathbb{R}^{J\times N}$; number of components $\mathcal{M}$; number of probes $\mathcal{P}$; temperature $T$; features dimension $J$;\textbf{}
\Statex \textbf{Output:} Distribution embedding $\mathbf{h}\in\mathbb{R}^{\mathcal{P}}$, GMM parameters $({\pi},{\mu},{\sigma}^2)$

\State \textbf{Soft responsibilities and instance importance:}
\For{$i=1$ \textbf{to} $N$}
    \State $ w_i \gets \mathrm{softmax}_i\!\left(\mathbf{f_w}(f_{x_i})\right) , \quad r_{i,k} \gets \mathrm{softmax}_k\!\left(\mathbf{f_r}(f_{x_i})\right), \quad k=1,\ldots,\mathcal{M}$
\EndFor

\State \textbf{Combine importance and responsibilities:}
\For{$i=1$ \textbf{to} $N$}
    \For{$k=1$ \textbf{to} $\mathcal{M}$}
        \State $a_{i,k} \gets w_i \cdot r_{i,k}$
    \EndFor
\EndFor


    
\State \textbf{Mixing coefficients and attention-weighted moments:}
\For{$k=1$ \textbf{to} $\mathcal{M}$}
    \State $\pi_k \gets \sum_{i=1}^{N} a_{i,k}
    \quad
    {\mu}_k \gets 
    \frac{\sum_{i=1}^{N} a_{i,k}{f}_{x_i}}
    {\sum_{i=1}^{N} a_{i,k}}
    \quad
    {\sigma}_k^2 \gets 
    \frac{\sum_{i=1}^{N} a_{i,k}(f_{x_i}-\boldsymbol{\mu}_k)^2}
    {\sum_{i=1}^{N} a_{i,k}}$
\EndFor
\State $\pi_k \gets \pi_k \big/ \left(\sum_{k'=1}^{\mathcal{M}} \pi_{k'} \right)$

\State \textbf{Distribution embedding via learned probes:}
\For{$p=1$ \textbf{to} $\mathcal{P}$}
    \State $h_p \gets \log\!\left(\sum_{k=1}^{\mathcal{M}} \pi_k\,
    \mathcal{N}\!\left({z}_p \mid {\mu}_k, {\sigma}_k^2\right)\right ) / (J \times T)$
\EndFor


\State \Return $\mathbf{h}$, $({\pi},{\mu},{\sigma}^2)$
\end{algorithmic}
\end{algorithm}

\section{Experimental Setup}
\subsection{Dataset}
We evaluate our method on a private transvaginal ultrasound (TVUS) dataset of cervical canal segmentations provided by clinicians. This study was approved by the Institutional Review Board of Corniche hospital (Protocol CH13032401). The cohort consists of 182 pregnant patients, including 44 preterm birth cases. Notably, all patients were prospectively classified as high-risk for PTB by the treating clinicians according to institutional clinical practice. This reflects the real clinical use of transvaginal ultrasound, where screening is not routinely performed in the general population but rather in women identified as being at high risk. This also makes the prediction task particularly challenging, as the elevated baseline risk reduces class separability between PTB and term cases. Each patient has between 1 and 43 ultrasound images (10$\pm$7 on average), acquired across one or multiple clinical visits. Most scans were acquired during the second trimester or near its boundaries (late first trimester and early third trimester). 


We additionally evaluate on a public lymph node histopathology MIL dataset \cite{oner2023distribution} for (i) binary classification (normal vs. metastasis) and (ii) regression of metastatic pixel percentage. The dataset consists of 933 training, 668 validation, and 736 test samples, of which 60\% are positive.

\subsection{Configurations, Implementation and Baselines}
We use a unified training protocol across pooling methods to ensure fair comparison. We experiment with five different pooling strategies, i.e., max, mean, attention \cite{ilse2018attention}, density \cite{oner2023distribution}, and ours (GMM), and investigate the prediction performance.  For PTB prediction, we use 5-fold cross-validation with patient-level splits and three random seeds per fold (15 runs). Models are trained for 80 epochs (batch size 3, learning rate $ 1 \times 10^{-4}$.). We use a U-Net \cite{ronneberger2015u}  as in \cite{wlodarczyk2020spontaneous}, training segmentation and classification jointly; the classification head aggregates images from the same patient to predict outcome and the segmentation mask predicts the cervical canal. For the lymph node dataset, we follow \cite{oner2023distribution} using a ResNet18 \cite{he2016deep} encoder with bag and batch size 32, repeating each experiment with three random initializations.

The number of mixture components $\mathcal{M}$, learnable probes $\mathcal{P}$, and temperature parameter $T$ are treated as dataset-specific hyperparameters since they are emprically optimized. 
For the preterm birth dataset, we set $(\mathcal{M}, \mathcal{P}, T) = (4, 96, 30)$, and for the lymph node metastasis dataset, $(\mathcal{M}, \mathcal{P}, T) = (10, 5, 30)$. 


\section{Results}

Table \ref{tab:preterm_results} compares different pooling strategies on the PTB dataset. The evaluation metrics are area under the precision–recall curve (PR-AUC) and area under the receiver operating characteristic curve (ROC-AUC) for classification, and the Dice score for segmentation. Given the class imbalance, PR-AUC is considered the primary metric.

Among MIL methods, max pooling, attention pooling, and GMM pooling achieve the highest PR-AUC scores, with differences within 0.01. To determine whether these differences are statistically meaningful, we conducted a paired $t$-test on PR-AUC across 15 paired experiments. The comparison between GMM and max pooling showed no significant difference, with a mean difference of $-0.005$ (95\% CI: $[-0.070,\,0.059]$, $p=0.873$), indicating comparable performance. Notably, GMM pooling exhibits lower standard deviation across runs, suggesting improved stability.


Table~\ref{tab:results_lymph} summarizes results on the lymph node benchmark. In contrast to the preterm dataset, max pooling shows reduced performance on this task, whereas GMM pooling achieves the best classification results ($0.91 \pm 0.01$ F1, $0.89 \pm 0.01$ ROC-AUC) and competitive regression performance ($0.18 \pm 0.02$ MAE), within 0.01 of the top regression method. These results suggest GMM pooling provides more consistent performance across tasks, achieving state-of-the-art results on the lymph node benchmark with lower variability.

\begin{figure*}[t]
    \centering
    
    \begin{subfigure}{0.32\textwidth}
        \centering
        \includegraphics[width=\linewidth]{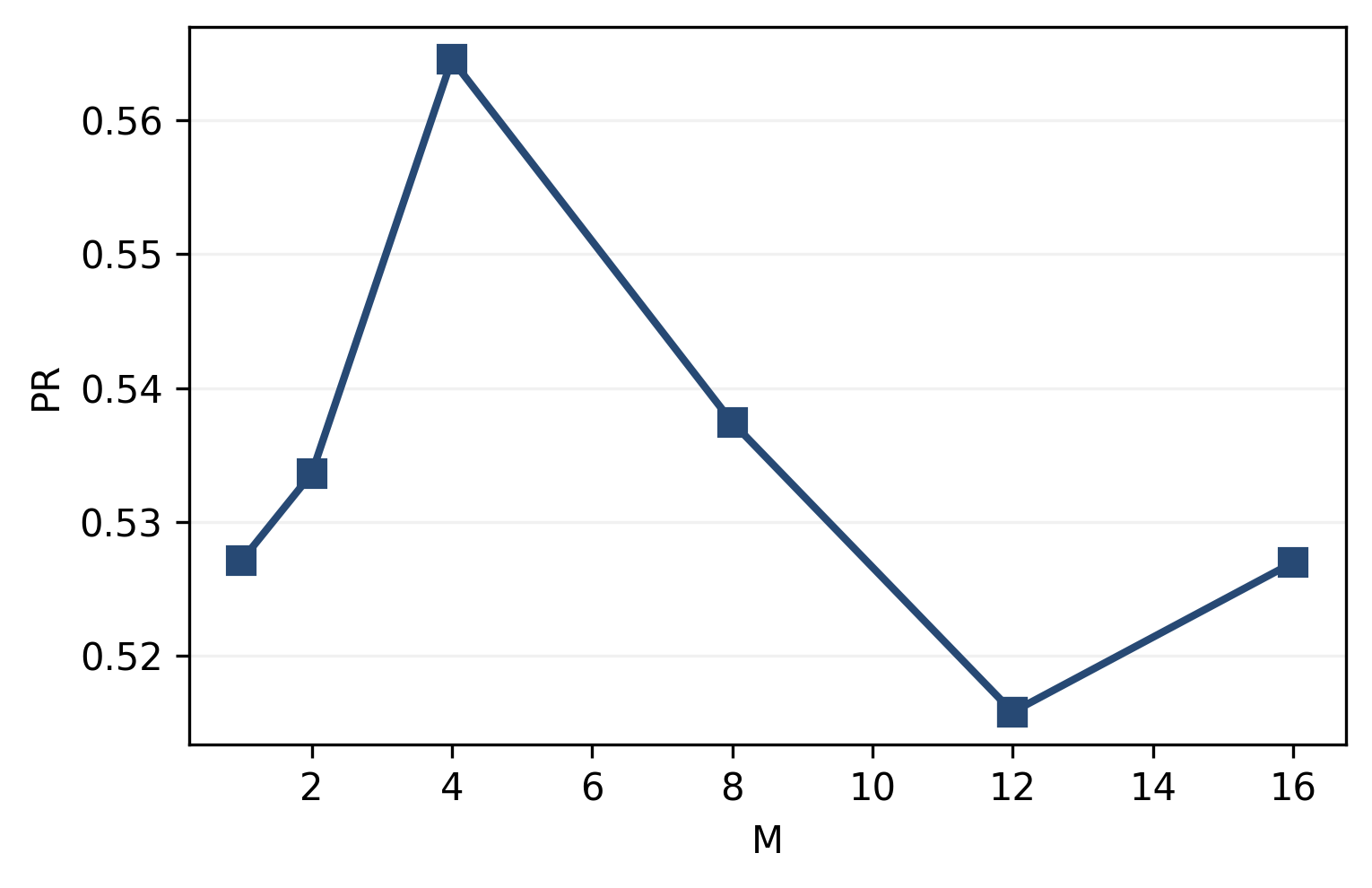}
        \caption{Effect of $M$ }
        \label{fig:ablation_M}
    \end{subfigure}
    \hfill
    \begin{subfigure}{0.32\textwidth}
        \centering
        \includegraphics[width=\linewidth]{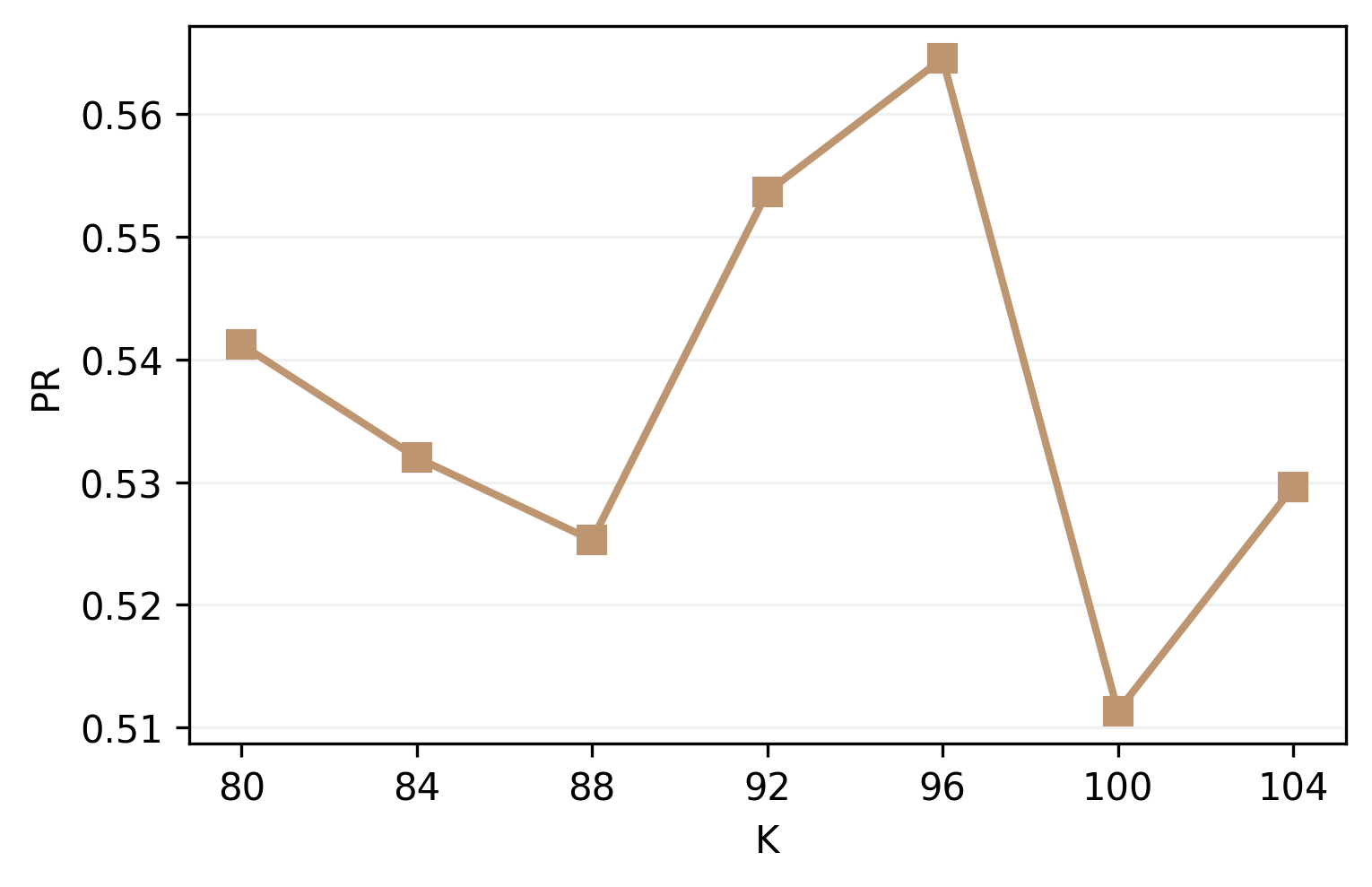}
        \caption{Effect of $P$}
        \label{fig:ablation_K}
    \end{subfigure}
    \hfill
    \begin{subfigure}{0.32\textwidth}
        \centering
        \includegraphics[width=\linewidth]{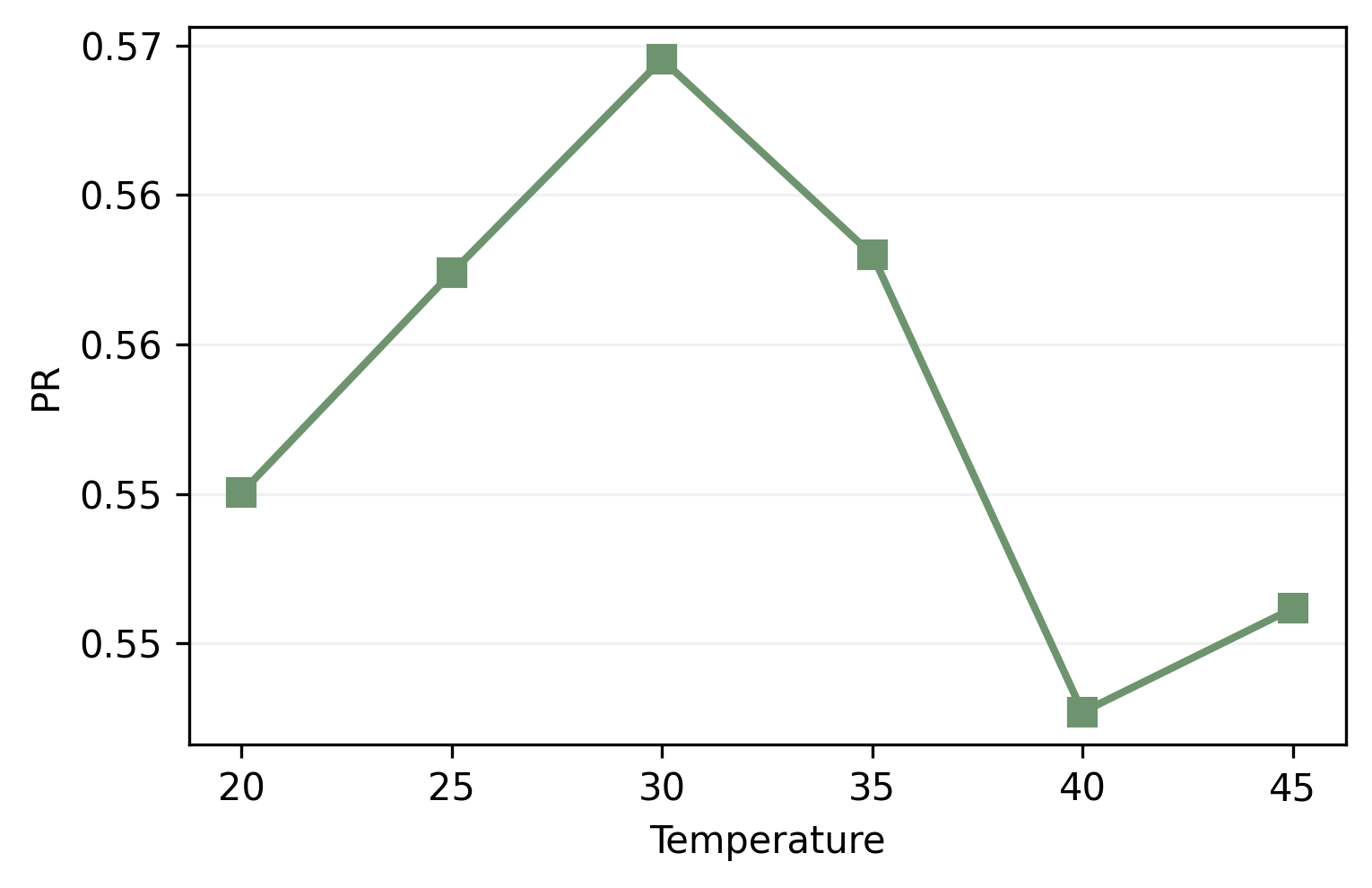}
        \caption{Effect of T }
        \label{fig:ablation_T}
    \end{subfigure}
    
    \caption{Ablations on GMM hyperparameter sensitivity on PTB (PR-AUC vs. $\mathcal{M}, \mathcal{P}, T)$.}
    \label{fig:ablation_all}
\end{figure*}

Figure \ref{fig:ablation_all} shows ablations on the hyperparameter sensitivity on the PTB dataset. Performance peaks at $\mathcal{M}=4$ and decreases for larger values. The probe count $\mathcal{P}$ achieves its best result at $\mathcal{P}=96$, while the temperature parameter $T$ performs optimally at $T=30$, with higher values reducing PR-AUC.

\begin{table*}[t]
\centering
\caption{Performance comparison on (left) preterm birth dataset and (right) lymph node dataset. Results are reported as mean $\pm$ std. Best results are in \textbf{bold}; second-best are \underline{underlined}. Segmentation performance is reported for completeness.}
\label{tab:two_results}

\begin{subtable}[t]{0.49\textwidth}
\centering
\caption{Preterm birth dataset.}
\label{tab:preterm_results}
\resizebox{\linewidth}{!}{%
\begin{tabular}{l c c | c}
\toprule
\rowcolor{gray!15}
 & \multicolumn{2}{c|}{\textbf{Classification}} & \textbf{Segmentation} \\
\rowcolor{gray!15}
\multirow{-2}{*}{\makecell[l]{Pooling\\Strategy}}
& \textbf{PR-AUC $\uparrow$}
& \textbf{ROC-AUC $\uparrow$}
& \textbf{Dice $\uparrow$} \\
\midrule
instance-based & 0.44$\pm$0.01 & 0.64$\pm$0.02 & 0.82\\
max           & \textbf{0.57$\pm$0.05} & \textbf{0.71$\pm$0.03} & 0.81 \\
mean          & 0.54$\pm$0.01 & 0.68$\pm$0.04 & \underline{0.82} \\
attention     & \underline{0.56$\pm$0.02} & 0.67$\pm$0.03 & \textbf{0.83} \\
density       & 0.54$\pm$0.03 & 0.68$\pm$0.02 & \underline{0.82} \\
\rowcolor{blue!15}
\textbf{GMM (ours)} & \underline{0.56$\pm$0.03} & \underline{0.69$\pm$0.03} & \underline{0.82} \\
\bottomrule
\end{tabular}}
\end{subtable}
\hfill
\begin{subtable}[t]{0.49\textwidth}
\centering
\caption{Lymph node dataset.}
\label{tab:results_lymph}
\resizebox{\linewidth}{!}{%
\begin{tabular}{l c c | c}
\toprule
\rowcolor{gray!15}
 & \multicolumn{2}{c|}{\textbf{Classification}} & \textbf{Regression} \\
\rowcolor{gray!15}
\multirow{-2}{*}{\makecell[l]{Pooling\\Strategy}}
& \textbf{F1 $\uparrow$}
& \textbf{ROC-AUC $\uparrow$}
& \textbf{MAE $\downarrow$} \\
\midrule
max       & 0.84$\pm$0.03 & 0.83$\pm$0.05 & 0.24$\pm$0.01 \\
mean      & 0.86$\pm$0.04 & 0.86$\pm$0.03 & 0.23$\pm$0.02 \\
attention & \underline{0.87$\pm$0.02} & \underline{0.88$\pm$0.02} & \underline{0.18$\pm$0.04} \\
density   & \underline{0.87$\pm$0.01} & 0.87$\pm$0.01 & \textbf{0.17$\pm$0.01} \\
\rowcolor{blue!15}
\textbf{GMM (ours)} & \textbf{0.91$\pm$0.01} & \textbf{0.89$\pm$0.01} & \underline{0.18$\pm$0.02} \\
\bottomrule
\end{tabular}}
\end{subtable}
\end{table*}

\section{Discussion}
Standard MIL pooling methods (max, mean, attention) summarize a variable-sized set of instance features into a single point estimate. While effective, this compression may discard informative intra-bag variability. In the context of transvaginal ultrasound, variability across images reflects differences in acquisition angle, anatomical visibility, and subtle cervical patterns. Modeling this variability explicitly can therefore be beneficial for patient-level risk prediction.

GMM pooling represents each bag as a learned mixture distribution rather than a single summary vector. On the PTB cohort, MIL substantially improves over instance-level training, confirming the benefit of leveraging all available images per patient. Although GMM pooling performs comparably to max and attention pooling in terms of PR-AUC, it exhibits lower variability across runs, suggesting improved stability. On the lymph node benchmark, GMM pooling achieves the strongest classification performance and competitive regression accuracy, indicating that distributional modeling generalizes across tasks.

The main trade-off of the approach is the introduction of distribution-specific hyperparameters (e.g., number of mixture components and probes), which require dataset-dependent tuning. Nevertheless, results across two distinct medical imaging tasks suggest that modeling feature distributions provides a robust and flexible alternative to point-estimate pooling.

\section{Conclusion}
We formulate PTB prediction from TVUS as a multiple instance learning problem to leverage all cervical images acquired per patient. We introduce GMM pooling to model within-bag feature distributions. On a private high-risk PTB cohort, MIL improves substantially over instance-based training, and GMM pooling achieves competitive PR-AUC with lower variability across runs. On a public lymph node metastasis benchmark, GMM pooling attains state-of-the-art classification performance and strong regression results, suggesting good cross-task generality. Future work will explore adaptive strategies for selecting mixture complexity, conduct more comprehensive component-wise ablation studies to better understand the contribution of individual design choices, and evaluate the approach on additional clinical cohorts to further assess generalizability.
\\ \\
\noindent\textbf{Disclosure of Interests.} The authors have no competing interests in the paper
as required by the publisher.

\bibliographystyle{splncs04}
\bibliography{ref.bib}








\end{document}